\documentclass[pdflatex,sn-mathphys]{sn-jnl}

\jyear{2021}%
\usepackage[utf8]{inputenc} 
\usepackage[T1]{fontenc} 
\usepackage[english]{babel}
\usepackage[inline]{trackchanges}
\usepackage[singlelinecheck=false,justification=justified]{caption}
\usepackage{graphicx}
\usepackage{subcaption}
\usepackage{hyperref}

\theoremstyle{thmstyleone}%
\theoremstyle{thmstyletwo}%

\theoremstyle{thmstylethree}%

\begin{document}

\title[DVS object counting]{Fast-moving object counting with an event camera}


\author{\fnm{Kamil} \sur{Bialik}} \email{kbialik@student.agh.edu.pl}
\author{\fnm{Marcin} \sur{Kowalczyk \href{https://orcid.org/0000-0002-4257-8877}{\includegraphics[width=10pt]{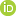}}}}  \email{kowalczyk@agh.edu.pl} 
\author{\fnm{Krzysztof} \sur{Błachut~\href{https://orcid.org/0000-0002-1071-335X}{\includegraphics[width=10pt]{orcid.png}}}} \email{kblachut@agh.edu.pl} 
\author*{\fnm{Tomasz} \sur{Kryjak}* \href{https://orcid.org/0000-0001-6798-4444}{\includegraphics[width=10pt]{orcid.png}}} \email{tomasz.kryjak@agh.edu.pl}   

\affil{\orgdiv{Embedded Vision Systems Group, Computer Vision Laboratory, Department of Automatic Control and Robotics}, \orgname{AGH University of Science and Technology}, \orgaddress{\street{Al. Mickiewicza 30}, \city{Krakow}, \postcode{30-059}, \country{Poland}}}


\abstract{
This paper proposes the use of an event camera as a~component of a~vision system that enables counting of fast-moving objects -- in this case, falling corn grains.
These type of cameras transmit information about the change in brightness of individual pixels and are characterised by low latency, no motion blur, correct operation in different lighting conditions, as well as very low power consumption.
The proposed counting algorithm processes events in real time.
The operation of the solution was demonstrated on a~stand consisting of a~chute with a~vibrating feeder, which allowed the number of grains falling to be adjusted.
The objective of the control system with a~PID controller was to maintain a~constant average number of falling objects.
The proposed solution was subjected to a~series of tests to determine the correctness of the developed method operation.
On their basis, the validity of using an event camera to count small, fast-moving objects and the associated wide range of potential industrial applications can be confirmed.}

\kewwords{event camera, dynamic vision sensor, high-speed object counting, item flow regulation, portioning, PID controller}



\maketitle

\section{Introduction}\label{sec:wprowadzenie}

Object counting has many practical applications.
It can be used in production lines as a~component of a~portioning and packaging system.
Examples include checking the number of tablets to be placed in packages or nuts that should be attached to the product being sold.
Such systems typically need to be able to count small objects that move at high speeds.
One possible solution to this problem is the use of a~vision system, i.e. the use of a~camera pointed at the objects to be counted.
Vision information usually allows for precise counting of moving objects. However, algorithms processing these data may return incorrect results when objects in the image occlude each other.
Another limitation of the vision solution is the low acquisition frequency, which causes blurring of the image when the speed of the counted objects is high. 
Reducing their speed is sometimes not possible or, in the case of a~production line, may involve a~loss of productivity.


Vision sensors that are good at observing fast-moving objects are event cameras, also referred to as DVS (Dynamic Vision Sensor).
Their principle of operation is inspired by the human eye. Instead of capturing an image for every fixed unit of time, each pixel operates independently of the rest and reacts continuously to a~change in the light intensity. If such a~change is detected, a~so-called event is generated, which consists of four pieces of information.
The first one is the timestamp. It identifies the moment at which the change occurred. In the latest sensors, its accuracy is \(1 \mu s\) \cite{gallego2022}. The next two components are the coordinates of the pixel in the matrix (\textit{x} and \textit{y} respectively). The last component is the polarisation. It determines whether the intensity of light falling on a~given pixel has increased or decreased.
Although the technology is relatively young (the first commercially available camera was unveiled in 2008 \cite{LiPoDe}), it is attracting considerable interest.
The leaders in the event camera industry are the corporations \textit{Samsung}, \textit{iniVation} and \textit{Prophesee}.

Event sensors have a~number of advantages over their traditional counterparts, which sequentially send pixels that belong to the entire image frame.
They have a~very high temporal resolution. Pixel brightness monitoring is fast, thanks to the analogue circuitry. 
On the contrary, the event readout itself is digital. The clock frequency used depends on the camera model, but is \(1 MHz\) for the latest sensors. This means that events are detected and time-stamped with microsecond resolution. This allows devices to capture very fast motion without blurring, which is typically a~problem with traditional frame cameras \cite{kim2016}.
DVS sensors are also characterised by quite low latency. Due to the independent operation of each pixel, there is no need to wait for the entire image frame to be collected and transmitted. Events can be sent as soon as a~change is detected. The latency is typically around \(100 \mu s\) for modern sensors.
As only brightness changes are sent, there is no redundant data. Power is only used to process pixels that have changed. This results in a~very low average power consumption of the photosensitive matrix (around \(50 mW\)) and a~much lower average power consumption of the entire system. This makes event camera solutions potentially very efficient and energy-saving.
These cameras are also characterised by a~much wider dynamic range of more than \(120 dB\). They significantly outperform even high-quality frame cameras, whose dynamic range is typically around \(60 dB\). This is achieved because the photoreceptors of the pixels operate on a~logarithmic scale and each one operates independently. Like human eyes, DVS can adapt to very dark and very bright scenes.


Event cameras, however, bring not only benefits but also challenges in terms of novel data processing and analysis methods.
First, a~system must be designed to process spatiotemporal visual information, which takes the form of a~sparse point cloud. The events have very high temporal resolution but are spatially dispersed. A~standard video stream, on the other hand, has low temporal resolution but dense spatial information. For this reason, the vision algorithms designed for images are not applicable to event data.


Second, typical image processing algorithms are designed to process the absolute values of pixels. DVS only provides binary information about a~change in brightness -- its increase or decrease. Such a~change can be the result of a~change in the illumination of the scene, the movement of objects, or the movement of the camera itself.
Furthermore, the event sensors currently available are characterised by a~relatively high level of noise \cite{padala2018}. 
This necessitates the use of filtering, which must take into account both the position and timestamps of the event stream.


A~technology competing with dynamic vision sensors in the application under consideration could be traditional high-frequency cameras. However, even the fastest frame cameras have a~lower temporal resolution, they only work in very good lighting conditions, and their prices are significantly higher than event cameras. In addition, the image from such cameras could be blurred in the case of objects moving at high speed, which in turn could necessitate the use of an even more expensive, higher-frequency solution.


This paper proposes an algorithm to count fast moving objects using an event camera.
Objects were assumed to be falling, i.e. moving in one direction along the sensor's field of view.
The operation of the solution was demonstrated on a~stand consisting of a~chute with a~vibrating feeder, which allowed the number of falling corn grains to be regulated.
The objective of the control system with a~PID (Proportional-Integral-Derivative) controller was to maintain a~constant average number of falling objects.
The proposed solution was subjected to a~series of tests to determine the correctness of the method developed.
The results in all tests were very close to the reference values. The automatic control system worked correctly, maintaining a~constant flow of grains over time.
The algorithm was run on the processor of a~typical PC in real time -- details are described at the end of Section \ref{ssec:algorytm}.
On the basis of the aforementioned features, the validity of using an event camera to count small, fast-moving objects and the associated wide range of potential industrial applications can be confirmed.


The reminder of this paper is organised as follows.
Section \ref{sec:literatura} presents the literature related to the topic of this article.
Section \ref{sec:metoda} is devoted to a~detailed description of the proposed object counting method together with the test stand used.
A~description of the experiments and the results obtained are included in Section \ref{sec:rezultaty}.
The last Section \ref{sec:podsumowanie} contains a~summary of the completed work together with an indication of possible directions for the development of the project.

\section{Previous work}\label{sec:literatura}


The problem of counting objects using vision data, due to its practical application, has been analysed many times in the scientific literature.
The first part of this section discusses work on counting objects of different types using classical vision sensors.
The second part, in turn, presents work on a~similar topic using event-based sensors for this purpose.

\subsection{Object counting}


The paper \cite{lee2017real} presented the design of a~smart camera that was capable of recognising and counting objects. It consisted of a~\textit{Basler USB 3.0} industrial camera, an \textit{Nvidia Tegra TX1} platform containing 256 CUDA (Compute Unified Device Architecture) cores and an \textit{ARM Cortex A57} quad-core processor. A~SURF (Speeded-Up Robust Features) feature point detector was used to describe the detected objects, which were logos of two car brands. They moved along closed tracks that simulated production lines in a~factory. A~disadvantage of the proposed solution was the need to limit the maximum speed of the objects. Another limitation was the low computation speed of the SURF detector. For images with a~resolution of $960 \times 600$ pixels, it could only be calculated four times per second.


The authors of article \cite{baygin2018image} proposed a~system for object counting. The main components of this solution were Otsu thresholding and Hough transformation. For the input image, a~conversion from RGB to HSV space was performed. For the S component, Gaussian blurring was performed, followed by Otsu thresholding. For the binary image, edges were determined using Sobel filters. The image with the detected edges was further subjected to a~Hough transform. Based on the result of the transform, detection and counting were performed. It worked independently of the type and colour of the counted elements. The application used a~global shutter camera with a~resolution of \(1280 \times 720\) pixels @ 59 fps (frames per second). The above algorithm was only executed when the counted objects were at the designated location. However, the article does not provide the processing time for a~single iteration of the described algorithm and does not provide information on the maximum speed of the objects.


Object counting is often used in the context of monitoring the number of people or vehicles moving through an observed area -- an example of such work is \cite{liu2017fast}.
The authors used the generation of a~background model using the Gaussian mixture models method. The background model was then subtracted from the recorded images. The result was thresholded. In this way, foreground objects were detected. Additionally, an omnidirectional search algorithm was used to count moving foreground objects.
The processing speed of the proposed algorithm was 40 frames per second for images with a~resolution of \(320 \times 240\) pixels, and the counting accuracy of the proposed method was 84.27\%.
Other works on counting people in specific spaces include, for example, \cite{lee2012} or \cite{saidon2021automatic}. The problem of traffic analysis has been addressed, for example, in the work \cite{liu2021}.


Other scientific papers related to object counting include \cite{wu2021}, describing a~potential application in the pharmaceutical industry.
The authors of this paper used the Hough transform to count objects, which were rectangular pill boxes.
The proposed solution allowed the vision system to operate in real time.

\subsection{Event camera}

Despite the large number of published articles related to event sensor data processing, a~general overview of which can be found, for example, in \cite{gallego2022}, only one dealing with object counting could be found.


In \cite{belbachir2010real}, the authors proposed a~system for counting cyclists and pedestrians using two event cameras. From the recorded data, the position of the events in a three-dimensional space (stereovision) was determined. Then these were grouped together. On the basis of the size of the groups and the time of their presence in the camera field of view, they were classified as a~pedestrian or a~cyclist. The effectiveness of the proposed classification system was 92\%. The event camera pointed from above at objects moving along the path. The proposed system was not tested for fast-moving objects and at high flow rates (more than 100 objects per minute).


An algorithm to count objects from event data has also been proposed by \textit{Prophesee}.
However, the method presented has not been described in the scientific literature, so information about it is only available in the documentation provided by the company. Objects are counted on horizontal lines.
They are assumed to move from top to bottom. It is necessary to select the appropriate parameters of the algorithm to make the counting efficient.
These parameters are: the distance of the camera from the objects, their average speed and minimum size, the polarity of the events used, and the positions of the aforementioned lines.
Initially, the event stream is filtered according to the set parameters (polarisation and activity filter). However, the way object counting itself works is not described.
It is known that each line acts as an independent object counter. The result is the maximum number of objects of all lines.

\section{The proposed approach}\label{sec:metoda}

\begin{figure}[!t]%
\centering
\captionsetup{justification=centering}
\includegraphics[width=0.65\textwidth]{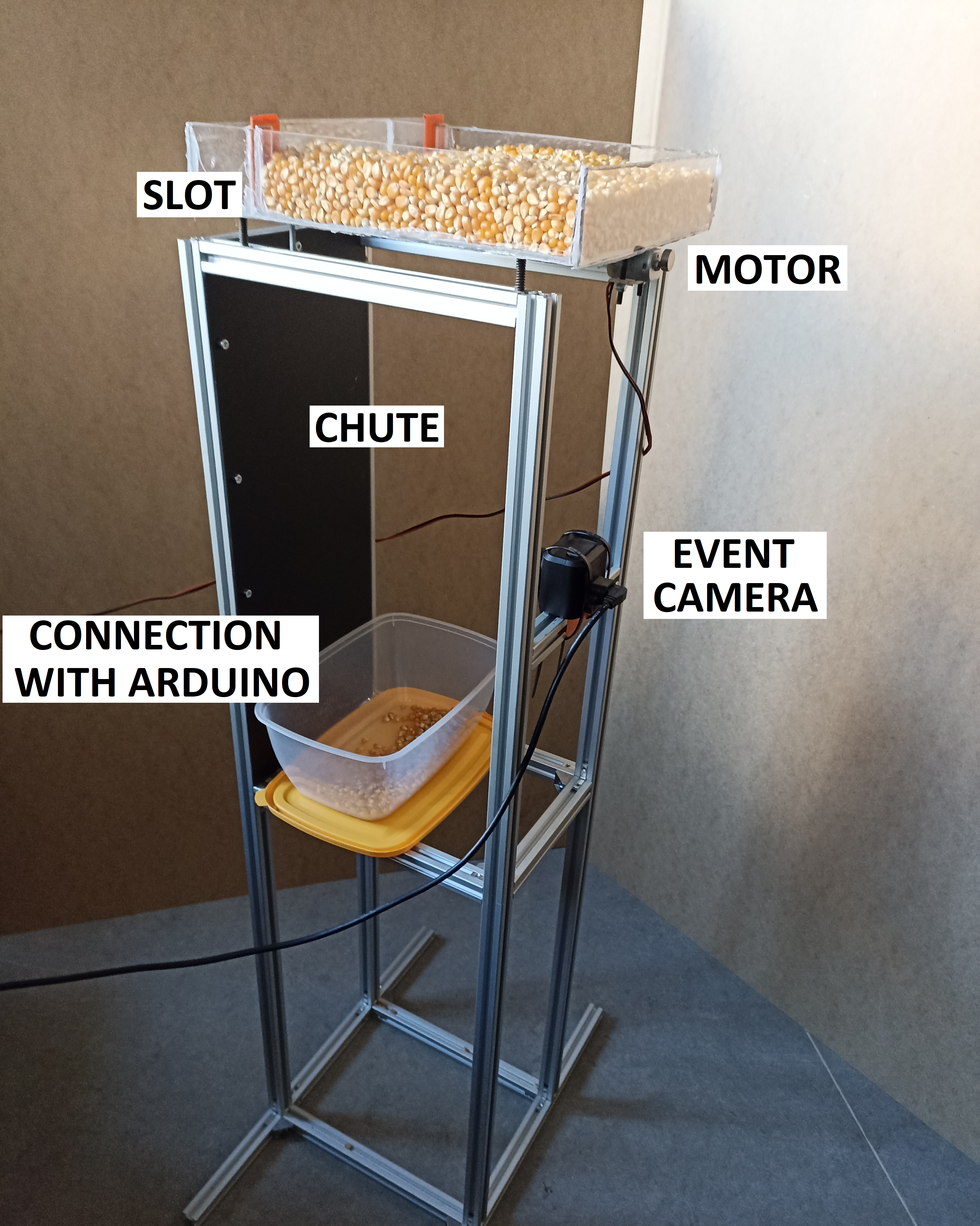}
\vspace{5pt}
\caption{Used test stand for object counting.}\label{fig:stanowisko}
\vspace{-5pt}
\end{figure}

Figure \ref{fig:stanowisko} shows the object counting test stand prepared for this project.
It consisted of a~vibrating feeder, an event camera and a~PC.
The corn grains were in a~feeder with an MT73 vibrating motor.
This motor was connected to an ESP8266 module with NodeMCU software, programmable with Arduino IDE (this module is omitted from the figure for clarity and is referred to as Arduino for simplicity in the rest of the article), controlling it in PWM (Pulse-Width Modulation) mode at a~frequency of 1 Hz.
At the other end of the feeder was a~small slot through which the grains moved, falling into the box below.
A~\textit{Prophesee} EVK1 event camera with HD resolution (\(1280 \times 720\) pixels) was mounted between the two, which was used to record changes in the brightness of the monitored area and, as a~result, to detect and count the falling grains.
The camera was connected via a~USB 3.0 bus to a~desktop PC, so that the system results could be observed on a~monitor connected to it.

\renewcommand{\figurename}{Figure}

\subsection{Vision algorithm}\label{ssec:algorytm}


The algorithm developed for object counting and motor control consisted of several components.
First, filters, provided by the manufacturer of the \textit{Prophesee} camera in the \textit{Metavision} software, were used to reduce the number of events processed.
The second operation was to create a~frame (image) from the events detected during a~specific time interval.
The off-the-shelf \textit{Metavision} software function was also used for this, and the event accumulation time was set to 2 ms, corresponding to image generation at 500 fps.
The pixels in the generated image were given values of 0 -- when there was no positive polarity event at a~given location during the 2 ms time period, and 255 -- when such an event was recorded at that location.


Then, functions from the OpenCV library -- \textit{findContours} and \textit{boundingRect} were used to determine the objects and their bounding boxes.
The next step was to compare the bounding boxes determined on the current and previous frame by calculating the value of the IoU (Intersection over Union) parameter, also known as the Jaccard index, according to Equation \eqref{eq:iou}.

\begin{equation}
\label{eq:iou}
IoU(A,B) = \frac{|A $\cap$ B|}{|A $\cup$ B|}
\end{equation}
\\where: A, B are two sets for which the IoU value is calculated.


If the value obtained was above the set threshold (experimentally, it was determined as 0.1), then the object was counted and information about it was recorded in the auxiliary table.
The use of such an approach allowed for the implementation of a~simplified object tracking, thanks to which they were distinguishable from each other (when several grains fell simultaneously), but also aimed to eliminate the potential problem of counting the same objects many times in successive frames.


For the object counting process itself, three so-called ``count lines'' (horizontal lines in the image) were used.
For the object matched between the current and previous frame, the centres of the bounding boxes (exactly the \textit{y} coordinates) were determined.
Due to the direction of movement of the corn grains (``down the image''), a~simple relationship was used -- on the previous frame the centre of the object should be above the specified horizontal line, and in the next frame below it.
On the basis of the tests carried out, it was decided to use three such lines to avoid situations where objects were not counted correctly.
The final result of counting objects on a~given frame was the maximum value of the three counting lines.

\begin{figure}[h]%
\centering
\captionsetup{justification=centering}
\includegraphics[width=0.65\textwidth]{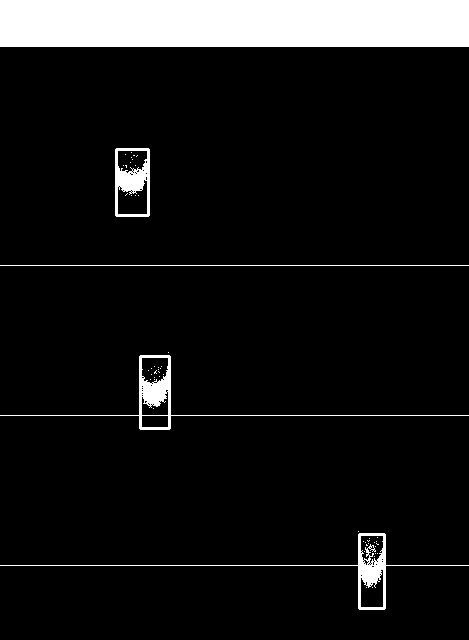}
\vspace{5pt}
\caption{Screenshot of the algorithm that counts falling objects.}\label{fig:algorithm}
\vspace{-5pt}
\end{figure}

Figure \ref{fig:algorithm} shows a~screenshot of the running algorithm. The count lines and the bounding boxes around the falling objects are visible.

The described vision algorithm a~was part of a~simple object flow control system.
Before it was run, the expected flow rate, i.e. the number of grains falling in one minute, was set.
The motor was supposed to vibrate the corn container in such a~way that the number of grains counted per unit of time corresponded to the set value.
Every 1 second, the error, understood as the difference between the number of objects counted and the expected value at that time, was calculated and controlled according to the formula of the discrete PID controller -- Equation \eqref{eq:pid}.

\begin{equation}
\label{eq:pid}
u_{n} = K_{p}e_{n} + K_{i} \sum_{i=1}^{n} e_{i} + K_{d}(e_{n}-e_{n-1})
\end{equation}
where:
\vspace{-0.25cm}
\begin{itemize}
    \item[] \hspace{0.05cm} $u_{n}$ -- control at time~n,
    \item[] \hspace{0.05cm} $K_{p}, K_{i}, K_{d}$ -- gains of the PID controller,
    \item[] \hspace{0.05cm} $e_{n}$ -- error at time~n.
\end{itemize}


The controller parameters were experimentally chosen -- $K_{p} = 2$ based on the occurrence of oscillations in the system, $K_{i} = 0.2$ so as to remove the residual error after a~few tens of seconds of operation, and $K_{d}$ was given a~small value of $0.1$ to better respond to the occurrence of congestions.
The resulting control value was then sent through the serial port to the Arduino, which provided signals to the motor in PWM mode, causing it to move for a~calculated fraction of a~second.
In addition, a~safety feature was added, shutting down the device in the event of a~congestion or emptying the grain container.


The object counting algorithm described above was written in Python and run on a~PC equipped with an Intel Core i7-7700K processor.
Due to the event filtering used and thus a~significant reduction in the number of events, the algorithm processed HD video data in real time. It should be noted that in the general case, the time is highly dependent on the number of events (including noise levels) and detected objects.

\section{Results obtained}\label{sec:rezultaty}


To evaluate the performance of the designed system, a~series of experiments was performed.
The first type of test involved stopping the system after counting a~predetermined number of grains.
Then the result of the operation was compared with a~reference value -- grains counted manually.
The results of the experiments are listed in Table \ref{tab:X_grains}.
Based on the analysis of the values, it can be concluded that the system worked correctly, occasionally omitting grains during counting.
Such situations occurred mainly when two grains fell close to each other (e.g., one partially occluded the other), causing the vision algorithm to treat them as a~single object.

\begin{table}[!t]
\begin{center}
\begin{minipage}{340pt}
\caption{Results of tests in which the system was stopped after counting a~certain number of objects. The values indicate the number of corn grains detected and counted.}\label{tab:X_grains}%
\centering
\begin{tabular}{@{}lccc@{}}
\toprule
Test & Reference value & Algorithm result & Error [pcs. (\%)] \\ 
\midrule
1.1 & 25  & 25  & 0  (0) \\
1.2 & 60  & 59  & -1 (1.7) \\
1.3 & 109 & 107 & -2 (1.8) \\
1.4 & 199 & 197 & -2 (1.0) \\
\botrule
\end{tabular}
\end{minipage}
\end{center}
\end{table}


The second type of experiment consisted of counting the grains that fell within 5 minutes of system operation at different values of the set flow rate.
Data from these tests were collected in Table \ref{tab:5_minutes}.
Again, there were some discrepancies between the results of the vision algorithm and the expected values at the output of the automatic control system, but they were negligible.

\begin{table}[!t]
\begin{center}
\begin{minipage}{340pt}
\caption{Results of experiments in which the automatic control system operated for 5 minutes for different values of the set flow rate (expressed in number of grains per minute). The values specify the number of corn grains.}\label{tab:5_minutes}%
\centering
\begin{tabular}{@{}lcccc@{}}
\toprule
Test & Set flow rate & Expected value & Algorithm result & Error [pcs. (\%)] \\ 
\midrule
2.1 & 50  & 250  & 243  & -7 (2.8) \\
2.2 & 200 & 1000 & 1002 & +2 (0.2) \\
2.3 & 300 & 1500 & 1499 & -1 (0.1) \\
\botrule
\end{tabular}
\end{minipage}
\end{center}
\end{table}


The purpose of the third type of test was to compare the performance of the designed vision algorithm with the sample object counting application from \textit{Prophesee}.
Several experiments were carried out, in which a~constant control signal (with different filling in each test) was given, recording events sequences of 1 minute length.
These were then passed as input to the algorithm proposed in this work and the application from \textit{Prophesee}.
The results are shown in Table \ref{tab:prophesee}.

\begin{table}[!t]
\begin{center}
\begin{minipage}{340pt}
\caption{Comparison of the results of the proposed algorithm for object counting with \textit{Prophesee}'s application. Values indicate the number of corn grains detected and counted.}\label{tab:prophesee}%
\centering
\begin{tabular}{@{}lccc@{}}
\toprule
Test & Reference value & \hspace{-5pt} Proposed algorithm/Error [pcs.(\%)] & \hspace{-5pt} \textit{Prophesee}/Error [pcs. (\%)] \\ 
\midrule
3.1 & 97  &  97/0 (0)  &  97/0  (0)  \\
3.2 & 115 & 115/0 (0)  & 116/+1 (0.9)  \\
3.3 & 191 & 190/-1 (0.5) & 194/+4 (2.1)  \\
3.4 & 129 & 130/+1 (0.8) & 131/+2 (1.6)  \\
3.5 & 199 & 199/0 (0)  & 201/+2 (1.0)  \\
3.6 & 173 & 173/0 (0)  & 175/+2 (1.2)  \\
\botrule
\end{tabular}
\end{minipage}
\end{center}
\end{table}


On the basis of the tests conducted, several conclusions can be drawn.
The proposed vision system works correctly, allowing the counting of falling small objects in real time.
The automatic control system also works properly, allowing the motor to be controlled in such a~way as to achieve the expected flow rate (a~set number of grains per unit of time).
In some experiments there are minor discrepancies between the results and the reference values, but as mentioned earlier, the main reason for this is the overlapping of objects in the image formed from the events.
On the other hand, the source of ``surplus counts'' are single situations in which a~given object is counted twice, such as when it is lost in the tracking process in several consecutive frames.
The aforementioned problems, however, occur rarely and less frequently than in the application from \textit{Prophesee}.

\section{Summary}\label{sec:podsumowanie}

In this this work, a~vision system equipped with an event camera was realised for counting fast-moving objects.
The implemented algorithm made it possible to analyse the events recorded by the camera in real time and, consequently, to detect and count objects.
The test stand prepared also consisted of a~chute equipped with a~vibrating feeder for objects (corn grains).


The experiments carried out confirmed the validity of the proposed approach -- the results obtained were very close to the reference values.
Only in some specific situations did the system not count all the grains correctly, such as when one grain was partially occluded by another.
The use of three counting lines significantly minimised the number of cases where individual grains were ``lost'' between consecutive frames.


Further work on the project could include several potential issues.
One of them is the use of a~set of several event cameras, monitoring the analysed space from different locations.
In this way, it should be possible to eliminate the aforementioned situations where objects are occluded or connected and difficult to count correctly.
Another issue could be the use of several mirrors, whose role would be similar to that of additional cameras -- to enable analysis of the space from multiple views, without generating such high costs as in the case of a~multi-camera system.
Yet another idea for further development of the project could be to count and recognise different types of objects falling down the chute.
The algorithm in its current or extended version can be also implemented on an embedded platform like Raspberry Pi, eGPU (e.g. NVidia Jetson) or SoC FPGA (e.g. Zynq from AMD Xilinx) to operate with much lower energy requirements.


The use of an event camera allows the detection and counting of small, fast-moving objects, making the proposed approach more versatile and, in many ways, better than solutions with traditional vision cameras.
The presented method can lay the foundation for the design of potential industrial applications, such as counting items on production lines or packing boxes.

\section*{Acknowledgement}\label{sec:podziekowania}
The work presented in this paper was supported by the programme ``Excellence initiative – research university'' for the AGH University of Science and Technology and was partly supported by the National Science Centre project no. 2021/41/N/ST6/03915 entitled ``Acceleration of processing event-based visual data with the use of heterogeneous, reprogrammable computing devices'' (second author).

\renewcommand\refname{References}


%

\end{document}